\documentclass[11pt,a4paper]{article}
\usepackage[hyperref]{acl2020}
\usepackage{times}
\usepackage{latexsym}
\usepackage{array}
\usepackage{tikz}
\usepackage{soul}
\usepackage{caption, floatrow}

\usepackage{times}
\usepackage{latexsym}
\usepackage{graphicx}
\usepackage{multirow}
\usepackage{todonotes}
\usepackage{amssymb}  
\usepackage{amsthm}
\usepackage{amsmath}
\usepackage{mathtools}
\newcommand{\Tref}[1]{Table~\ref{#1}}
\newcommand{\Eref}[1]{Eq.~(\ref{#1})}
\newcommand{\Fref}[1]{Figure~\ref{#1}}

\newcommand{\Sref}[1]{Section~\ref{#1}}
\usepackage{arydshln}
\usepackage{algorithm}
\usepackage{algorithmic}
\usepackage{graphicx}
\usepackage{booktabs}
\usepackage{calrsfs}
\usepackage{array}
\DeclareMathAlphabet{\mathcal}{OMS}{cmsy}{m}{n}

\newcolumntype{L}[1]{>{\raggedright\let\newline\\\arraybackslash\hspace{0pt}}m{#1}}
\newcolumntype{C}[1]{>{\centering\let\newline\\\arraybackslash\hspace{0pt}}m{#1}}
\newcolumntype{R}[1]{>{\raggedleft\let\newline\\\arraybackslash\hspace{0pt}}m{#1}}

\newcolumntype{?}{!{\vrule width 1pt}}

\usepackage{microtype}
\usepackage{subfiles}

\aclfinalcopy 

\title{Logic-Guided Data Augmentation and Regularization \\ for Consistent  Question Answering }

\author{Akari Asai$^{\dagger}$ \and Hannaneh Hajishirzi$^{\dagger	\ddagger}$\\
  ${\dagger}$University of Washington~~$\ddagger$Allen Institute for AI\\
  \texttt{\{akari, hannaneh\}@cs.washington.edu} 
  }

\begin{document}
\maketitle
\begin{abstract}
Many natural language questions require qualitative, quantitative or logical comparisons between two entities or events. 
This paper addresses the problem of improving  the accuracy and consistency of responses to comparison questions by integrating logic rules and neural models.  
Our method leverages logical and linguistic knowledge to augment labeled training data and then uses a consistency-based regularizer to train the model. 
Improving the global consistency of predictions, our approach achieves large improvements over previous methods in a variety of question answering (QA) tasks including multiple-choice qualitative reasoning, cause-effect reasoning, and extractive machine reading comprehension. In particular, our method significantly improves the performance of RoBERTa-based models by 1-5\% across datasets. We advance state of the art by around 5-8\% on WIQA and QuaRel and reduce consistency violations by 58\% on HotpotQA.
We further demonstrate that our approach can learn effectively from limited data.\footnote{Our code and data is available at \url{https://github.com/AkariAsai/logic_guided_qa}.}
\end{abstract}

\section{Introduction}
Comparison-type questions \cite{Tandon2019WIQAAD,tafjord2019quarel,yang-etal-2018-hotpotqa} ask about relationships between properties of entities or events such as cause-effect, qualitative or quantitative reasoning. 
To create comparison questions that require inferential knowledge and reasoning ability, annotators need to understand context presented in multiple paragraphs or carefully ground a question to the given situation. This makes it challenging to annotate a large number of comparison questions.
Most current datasets on comparison questions are much smaller than standard machine reading comprehension (MRC) datasets~\cite{rajpurkar-etal-2016-squad,joshi-etal-2017-triviaqa}. This poses new challenges to standard models, which are known to  exploit statistical patterns or annotation artifacts in these datasets~\cite{sugawara-etal-2018-makes,min2019compositional}. 
\begin{figure}
  \includegraphics[width=\linewidth]{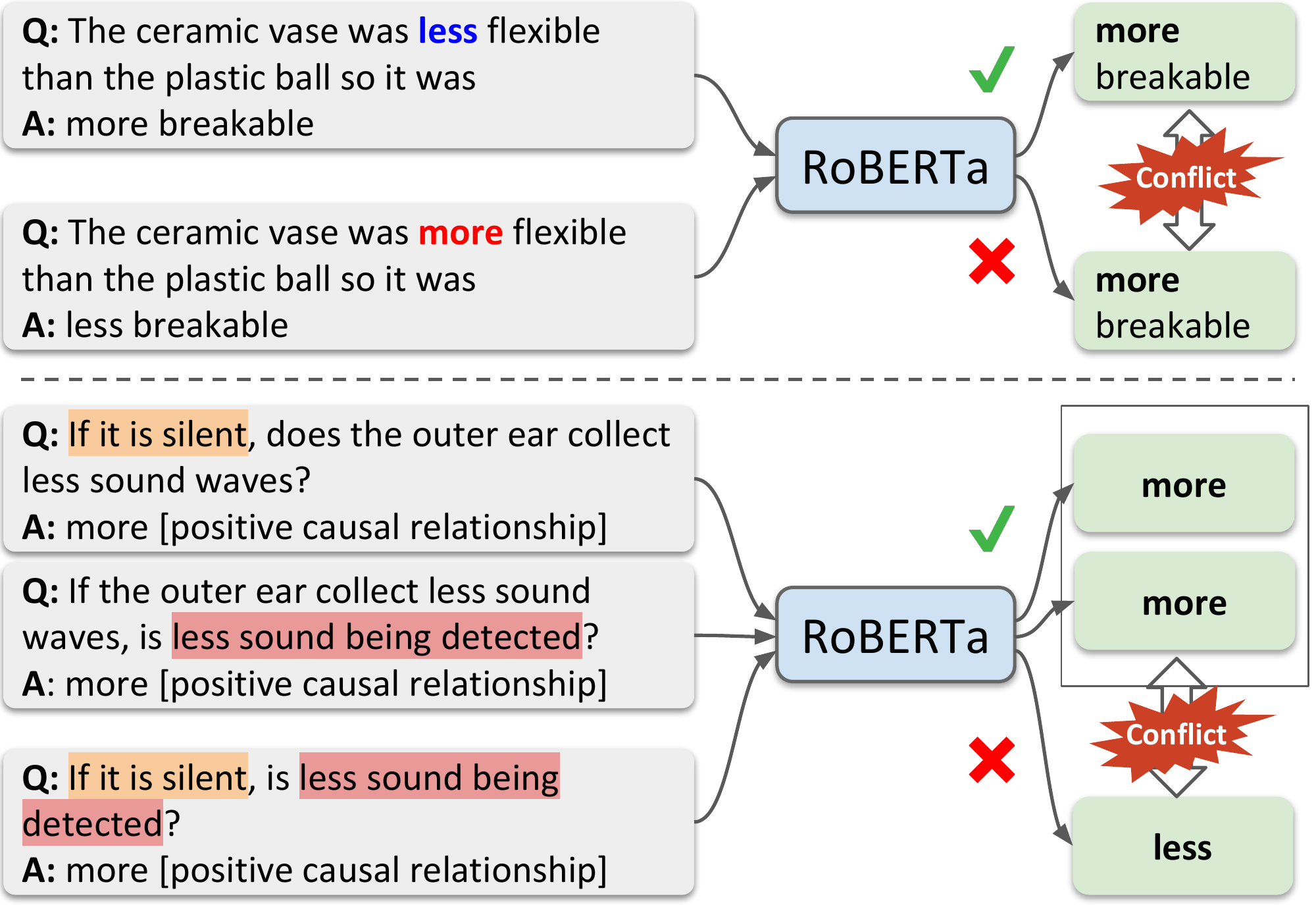}
  \caption{Inconsistent predictions by RoBERTa. Top row shows an example of symmetric inconsistency and the second row shows an example of transitive inconsistency. The examples are partially modified. }
  \label{tab:examples_trans_consistency}
\end{figure}
Importantly, state-of-the-art models show inconsistent comparison predictions as shown in \Fref{tab:examples_trans_consistency}..
Improving the consistency of predictions has been previously studied in natural language inference (NLI) tasks~\cite{minervini-riedel-2018-adversarially,li-etal-logical-nli-2019}, but has not been addressed in QA. 

In this paper, we address the  task of producing globally consistent and accurate predictions for comparison questions leveraging logical and symbolic knowledge for data augmentation and training regularization.
Our data augmentation uses a set of logical and linguistic knowledge to develop additional consistent labeled training data. 
Subsequently, our method uses symbolic logic to incorporate consistency regularization for additional supervision signal beyond inductive bias given by data augmentation. Our method generalizes previous consistency-promoting methods for NLI tasks~\cite{minervini-riedel-2018-adversarially,li-etal-logical-nli-2019} to adapt to substantially different question formats.

Our experiments show significant improvement over the state of the art on a variety of QA tasks: a classification-based causal reasoning QA, a multiple choice QA for qualitative reasoning and an extractive MRC task with comparisons between entities. 
Notably, our data augmentation and consistency constrained training regularization improves performance of RoBERTa-based models~\cite{liu2019roberta} by 1.0\%, 5.0\% and 2.5\% on WIQA, QuaRel and HotpotQA. Our approach advances the state-of-the-art results on WIQA and QuaRel with 4.7 and 8.4\% absolute accuracy improvement, respectively, reducing inconsistent predictions.
We further demonstrate that our approach can learn effectively from limited labeled data: given only 20\% of the original labeled data, our method achieves performance on par with a competitive baseline learned with the full labeled data.


\section{Related Work}Data augmentation has been explored in a variety of tasks and domains~\cite{krizhevsky2009learning,Cubuk_2019_CVPR,park2019specaugment}. 
In NLP, using back-translation~\cite{wei2018fast} or dictionary based word replacement~\cite{zhang2015character} has been studied. 
Most relevant to our work, \citet{kang-etal-2018-adventure} study NLI-specific logic and knowledge-based data augmentation. 
Concurrent to our work, \citet{gokhale2020vqa} study visual QA models' ability to answer logically composed questions, and show the effectiveness of logic-guided data augmentation. 
Our data augmentation does not rely on task-specific assumptions, and can be adapted to different formats of QA task. We further leverage consistency-promoting regularization, which gives improvements in accuracy and consistency. 

Improving prediction consistency via training regularization has been studied in NLI tasks. 
\citet{minervini-riedel-2018-adversarially} present model-dependent first-order logic guided adversarial example generation and regularization.
\citet{li-etal-logical-nli-2019} introduce consistency-based regularization incorporating the first-order logic rules. 
Previous approach is model-dependent or relies on NLI-specific rules, while our method is model-agnostic and is more generally applicable by combining it with data augmentation.

Regularizing loss to penalize violations of structural constraints in models' output has been also studied in previous work on constraint satisfaction in structured learning~\cite{lee2019gradient,ganchev2010posterior}.
Our work regularizes models to produce globally consistent predictions among augmented data following logical constraints, while those studies incorporates
structured prediction models following linguistics rules.



\section{Method}We present the components of our QA method: first-order logic guided data augmentation (\Sref{sec:background_knowledge} and \Sref{sec:augmentation}), and consistency-based regularization~(\Sref{sec:consistency_loss}). 

\subsection{Consistent Question Answering}
\label{sec:background_knowledge}
For globally consistent predictions in QA, we require responses to follow two important general logical rules: {\it symmetric consistency} and {\it transitive consistency}, which are illustrated in \Fref{tab:examples_trans_consistency} and are formally described below.

Let $q, p, a$ be a question, a paragraph and an answer predicted by a model. 
$\mathbf{A}$ is a set of answer candidates. 
Each element of $\mathbf{A}$ can be a span in $p$, a class category, or an arbitrary answer choice.
$X = \{q, p, a\}$ represents a logic atom. 

\begin{table*}[t!]
\begin{center}
{\small
\begin{tabular}{c|c|c|c}
\toprule
  & WIQA & QuaRel & HotpotQA \\
  & \citep{Tandon2019WIQAAD} & \citep{tafjord2019quarel} &
  \citep{yang-etal-2018-hotpotqa} \\\hdashline
 reasoning & Causal Reasoning & Qualitative Reasoning & Qualitative Comparison of  entities \\
  format & classification & multiple choice & span extraction \\
\midrule
\multirow{4}{*}{$p$} &\multirow{4}{*}{\parbox{5cm}{The rain seeps into the wood surface. When rain evaporates it leaves the wood. It takes the finish of the wood with it. The wood begins to lose it's luster.}}  & \multirow{4}{*}{\parbox{3.5cm}{Supposed you were standing on the planet Earth and Mercury. When you look up in the sky and see the sun, }}  &  \multirow{4}{*}{\parbox{4.5cm}{Golf Magazine is a monthly golf magazine owned by Time Inc. El Nuevo Cojo Ilustrado is an American Spanish language magazine.}}  \\
&&&\\
&&&\\
&&&\\\hline
\multirow{3}{*}{$q$} &\multirow{3}{*}{\parbox{5cm}{$q_1$:If a tsunami happens, will {\bf wood be more moist}?, $q_2$: If {\bf wood is more moist}, is more weathering occurring?}}  & \multirow{3}{*}{\parbox{3.5cm}{Which planet would the sun appear \textcolor{red}{\bf larger}?}}  &  \multirow{3}{*}{\parbox{4.5cm}{El Nuevo Cojo and Golf Magazine, which one is owned by Time Inc?}}  \\
&&& \\
&&& \\\hline
$\mathbf{A}$ & \{more, less, no effects\} & \{Mercury, Earth\}& \{Golf Magazine, El Nuevo Cojo\}  \\\hline
$a^{*}$ & $a_1^*$ : more, $a_2^*$ : more &Mercury & Golf Magazine \\\hline
\multirow{2}{*}{$q_{aug}$} & \multirow{2}{*}{\parbox{5cm}{{If a tsunami happens, is more weathering occurring?}}} & \multirow{2}{*}{\parbox{3.5cm}{Which planet would the sun appear \textcolor{blue}{\bf smaller}?}} & \multirow{2}{*}{\parbox{4.5cm}{Which one is \hl{\bf not} owned by Time Inc, Golf Magazine El Nuevo Cojo?}}  \\
&&& \\\hline
$a_{aug}^{*}$ & more & Earth & El Nuevo Cojo\\\bottomrule
\end{tabular}
}
\end{center}
\caption{\label{tab:QA_example} An augmented transitive example for WIQA, and symmetric examples for QuaRel and HotpotQA. 
We partially modify paragraphs and questions. 
The bold characters denote a shared event connecting two questions. 
The parts written in red or blue denote antonyms, and highlighted text is negation added by our data augmentation.
}
\end{table*}

\paragraph{Symmetric consistency}
In a comparison question, small surface variations such as replacing words with their antonyms can reverse the answer, while keeping the overall semantics of the question as before. 
We define symmetry of questions in the context of QA as follows: $(q, p, a^{*})\leftrightarrow (q_{sym}, p, a_{sym}^{*})$, where $q$ and $q_{sym}$ are antonyms of each other, and $a^{*}_{sym}$ is the opposite of the ground-truth answer $a^{*}$ in $\mathbf{A}$.   
For example, the two questions in the first row of \Fref{tab:examples_trans_consistency} are symmetric pairs. 
We define the symmetric consistency of predictions in QA as the following logic rule: 
\vspace{-0.2cm}
\begin{equation}\label{eq:inductive}
\vspace{-0.1cm}
    (q, p, a) \rightarrow (q_{sym}, p, a_{sym}),
\end{equation}
which indicates a system should predict $a_{sym}$ given $(q_{sym}, p)$, if it predicts $a$ for $(q, p)$.

\paragraph{Transitive consistency.}
Transitive inference between three predicates $A,B,C$ is represented as: $A \rightarrow B \wedge B \rightarrow C$ then $A \rightarrow C$ ~\cite{gazes2012cognitive}. 
In the context of QA, the transitive examples are mainly for causal reasoning questions that inquire about the effect $e$ given the cause $c$.
The second row of Figure~\ref{tab:examples_trans_consistency} shows an example where transitive consistency is violated.
For two questions $q_1$ and $q_2$ in which  the effect of $q_1$ ($=e_1$) is equal to the cause of $q_2$ ($=c_2$), we define the transitive consistency of predictions as follows:
\vspace{-0.2cm}
\begin{equation}
\vspace{-0.08cm}
    (q_1, p, a_1) \wedge (q_2, p, a_2) \rightarrow (q_{trans}, p, a_{trans}).
\end{equation}

\subsection{Logic-guided Data Augmentation}
Given a set of training examples $X$ in the form of $(q, p, a^{*})$, we automatically generate additional examples $X_{aug} = \{q_{aug}, p, a^*_{aug}\}$ using symmetry and transitivity logical rules. The goal is to augment the training data so that symmetric and transitive examples are observed during training. 
We provide some augmented examples in \Tref{tab:QA_example}.
\label{sec:augmentation}

\paragraph{Augmenting symmetric examples}
To create a symmetric question, we convert a question into an opposite one using the following operations:  (a) replace words with their antonyms, (b) add, or  (c) remove  words. 
For (a), we  select top frequent adjectives or verbs with polarity (e.g., {\it smaller, increases}) from training corpora, and expert annotators write antonyms for each of the frequent words (we denote this small dictionary as $\mathbf{D}$). More details can be seen in Appendix A.
For (b) and (c), we add negation words or remove negation words (e.g., {\it not}). 
For all of the questions in training data, if a question includes a word in $\mathbf{D}$ for the operation (a), or matches a template (e.g., \texttt{which * is} $\leftrightarrow$ \texttt{which * is not}) for operations (b) and (c), we apply the operation to generate $q_{sym}$.\footnote{We observe that (b)(c) are less effective than (a) in WIQA or QuaRel, while especially (b) contributes to the performance improvements on HotpotQA as much as (a) does. }
We obtain $a_{sym}^{*}$ by re-labeling the answer $a^*$ to its opposite answer choice in $\mathbf{A}$ (see Appendix B).

\paragraph{Augmenting transitive examples}
We first find a pair of two cause-effect questions $X_1 = (q_1, p, a_1^{*})$ and $X_2 = (q_2, p, a_2^{*})$, whose $q_1$ and $q_2$ consist of $(c_1, e_1)$ and $(c_2, e_2)$, where $e_1 = c_2$ holds. When $a_1^{*}$ is a {\it positive causal relationship},
we create a new example $X_{trans} = (q_3,p, a_2^{*})$ for 
$q_3 = (c_1, e_2)$. 
\begin{table*}[t!]
    \centering
    \small
\small
\begin{tabular}{ l | c  c c  c| c ? c  c  c| c ? c c |c}\toprule 
Dataset &  \multicolumn{5}{c}{WIQA}&  \multicolumn{4}{c}{QuaRel} & \multicolumn{3}{c}{HotpotQA} \\
& \multicolumn{3}{c}{Dev} & Test & $v$ (\%) &  \multicolumn{2}{c}{Dev} & Test  & $v$ (\%) & \multicolumn{2}{c}{Dev} & $v$ (\%)\\ \hline
x\% data & 20\% & 40\% & 100 \%  &100\% & 100\% & 20\% & 100\% & 100\% & 100\%& 20\% & 100 \% & 100 \% \\
(\# of $X$) & (6k) & (12k) & (30k) & (30k) &(30k) & (0.4k) & (2k) & (2k)  & (2k)  & (18k) & (90k) &  (90k) \\
\midrule
SOTA& -- & -- & -- &73.8 & -- & -- &  --  & 76.6 & -- & -- & --& -- \\ \hline
RoBERTa & 61.1 &  74.1& 74.9 &  77.5 & 12.0 & 56.4 & 81.1 & 80.0 & 19.2 & 71.0 & 75.5 & 65.2 \\\hline
\texttt{DA} &  72.1  &75.5 &  76.3& 78.3 & 6.0  & 69.3 &84.5 &  84.7 & 13.3 & \bf 73.1 & \bf 78.0 & \bf 6.3\\
\texttt{DA} + \texttt{Reg} &  \bf 73.9 & \bf 76.1 & \bf 77.0 & \bf 78.5 & \bf 5.8 & \bf 70.9 &\bf 85.1 & \bf 85.0 & \bf 10.3 & 71.9 & 76.9 & 7.2 \\
\bottomrule
\end{tabular}
\caption{{\bf WIQA, QuaRel and HotpotQA results}:we report test and development accuracy (\%) for WIQA and QuaRel and development F1 for HotpotQA. 
\texttt{DA} and \texttt{Reg} denote data augmentation and consistency regularization. 
``SOTA'' is \citet{Tandon2019WIQAAD} for WIQA and  \citet{mitra2019generate} for QuaRel. 
$v$ presents violations of consistency.
}\label{tab:result_wiqa}
\end{table*}
\paragraph{Sampling augmented data}
Adding all consistent examples may change the data distribution from the original one, which may lead to a deterioration in performance~\cite{xie2019adversarial}. 
One can select the data based on a model's prediction inconsistencies~\cite{minervini-riedel-2018-adversarially} or randomly sample at each epoch~\cite{kang-etal-2018-adventure}. 
In this work, we randomly sample augmented data at the beginning of training, and use the same examples for all epochs during training. Despite its simplicity, this yields competitive or even better performance than other sampling strategies.\footnote{We do not add $X_{aug}$ if the same pair has already exist.} 

\subsection{Logic-guided Consistency Regularization}
\label{sec:consistency_loss}
We regularize the learning objective (task loss, $\mathcal{L}_{task}$) with a regularization term that promotes consistency of predictions (consistency loss, $\mathcal{L}_{cons}$). 
\vspace{-0.2cm}
\begin{equation}\label{eq:multi_task_loss}
\vspace{-0.2cm}
    \mathcal{L} = \mathcal{L}_{task}(X) + \mathcal{L}_{cons}(X, X_{aug}). 
\end{equation}
The first term $\mathcal{L}_{task}$ penalizes making incorrect predictions. The second term $\mathcal{L}_{cons}$\footnote{We mask the $\mathcal{L}_{cons}$ for the examples without symmetric or transitive consistent examples.} penalizes making predictions that violate symmetric and transitive logical rules as follows: \vspace{-0.2cm}
\begin{equation}\label{eq:multi_task_loss_consistency}
\vspace{-0.2cm}
    \mathcal{L}_{cons} =  \lambda_{sym} \mathcal{L}_{sym} +\lambda_{trans} \mathcal{L}_{trans},
\end{equation}
where $\lambda_{sym}$ and  $\lambda_{trans}$ are weighting scalars to balance the two consistency-promoting objectives. 

Previous studies focusing on NLI consistency~\cite{li-etal-logical-nli-2019} calculate the prediction inconsistency between a pair of examples by swapping the premise and the hypothesis, which cannot be directly applied to QA tasks.
Instead, our method leverages consistency with data augmentation to create paired examples based on general logic rules. 
This enables the application of consistency regularization to a variety of QA tasks. 
\vspace{-0.3cm} 
\paragraph{Inconsistency losses}
The loss computes the dissimilarity between the predicted probability for the original labeled answer and the one for the augmented data defined as follows: 
\vspace{-0.2cm}
\begin{equation}
\vspace{-0.2cm}
    \mathcal{L}_{sym} = |{\rm log}~p(a | q, p) -  {\rm log}~p(a_{aug} | q_{aug}, p)|.
\end{equation}

Likewise, for transitive loss, we use absolute loss
with the product T-norm which projects a logical conjunction operation $(q_1, p, a_1) \wedge (q_2, c, a_2)$ to a product of probabilities of two operations, $p(a_1|q_1, p)p(a_2|q_2, p)$, following~\citet{li-etal-logical-nli-2019}. 
We calculate a transitive consistency loss as:
\vspace{-0.2cm}
\begin{equation} \label{eq:transitiveity}
\vspace{-0.2cm}
\begin{split}
\mathcal{L}_{trans} & = |{\rm log}~p(a_1 | q_1, p) + {\rm log}~p(a_2 | q_2, p) - \\
 & {\rm log}~p(a_{trans} | q_{trans}, p)|. \notag
\end{split}
\end{equation}

\vspace{-0.3cm} 
\paragraph{Annealing}
The model's predictions may not be accurate enough at the beginning of training for consistency regularization to be effective. 
We perform annealing~\cite{kirkpatrick1983optimization,li-etal-logical-nli-2019,du-etal-2019-consistent}. 
We first set $\lambda_{\{sym, trans\} }=0$ in \Eref{eq:multi_task_loss_consistency} and train a model for $\tau$ epochs, and then train it with the full objective. 


\section{Experiments}
\vspace{-.2cm}
\paragraph{Datasets and experimental settings}
We experiment on three QA datasets: WIQA~\cite{Tandon2019WIQAAD}, QuaRel~\cite{tafjord2019quarel} and HotpotQA (oracle, comparison questions\footnote{We train models on both bridge and comparison questions, and evaluate them on extractive comparison questions only.})~\cite{yang-etal-2018-hotpotqa}. 
As shown in Table~\ref{tab:QA_example}, these three datasets are substantially different from each other in terms of required reasoning ability and task format.
In WIQA, there are 3,238 symmetric examples and 4,287 transitive examples, while 50,732 symmetric pairs and 1,609 transitive triples are missed from the original training data.  HotpotQA and QuaRel do not have any training pairs requiring consistency. Our method randomly samples 50, 80, 90\% of the augmented data for WIQA, QuaRel and HotpotQA, resulting in 24,715/836/3,538 newly created training examples for those datasets, respectively.

\begin{table}[t!]
    \centering
    \small
\begin{tabular}{l|c c | c  c  }\toprule
 & \multicolumn{2}{c}{WIQA} &\multicolumn{2}{c}{QuaRel}\\\midrule
metric & acc & $v$ (\%) & acc  & $v$ (\%) \\\midrule
\texttt{DA} (logic) + \texttt{Reg} & 77.0 & 5.8 & 85.1 & 10.3\\ 
\texttt{DA}~(logic) & 76.3 & 6.0 & 84.5 & 13.5 \\ 
\texttt{DA}~(standard) & 75.2 & 12.3 & 83.3 & 14.5 \\ 
\texttt{Reg} & 75.8 & 11.4 & -- & --   \\\hdashline
Baseline & 74.9 & 12.0  & 81.1 & 19.2 \\
\bottomrule
\end{tabular}
\caption{Ablation studies of data augmentation on WIQA and QuaRel development dataset. }\label{tab:ablation_analysis}
\end{table}
\begin{table*}[t!]
\begin{center}
{\scriptsize
\begin{tabular}{C{0.85cm} | L{9cm}  |C{1.25cm} C{1.25cm} C{1.25cm} }
\toprule
& WIQA Input & RoBERTa & \texttt{DA} &\texttt{DA+Reg} \\
\midrule
$p$& Sound enters the ears of a person. The sound hits a drum that is inside the ears. & & & \\
$q$& If the person has his ears more protected, will less sound be detected? [$a^*$: \textcolor{red}{More}] & \textcolor{red}{More (0.79)} & \textcolor{red}{More (0.93)} & \textcolor{red}{More (0.93)}\\
$q_{sym}$ & If the person has his ears less protected, will less sound be detected? [$a^{sym}*$: \textcolor{blue}{Less}] & \textcolor{red}{More (0.87)} & \textcolor{red}{More (0.72)} & \textcolor{blue}{Less (0.89)}\\\hline
$p$& Squirrels try to eat as much as possible. Squirrel gains weight. \\
$q_1$&If the weather has a lot of snow, cannot squirrels eat as much as possible? [$a^*_1$: \textcolor{red}{More}]& \textcolor{blue}{Less (0.75)} & \textcolor{red}{More (0.48)} & \textcolor{red}{More (0.94)}\\
$q_2$&If squirrels cannot eat as much as possible, will not the squirrels gain weight? [$a^*_2$: \textcolor{red}{More}] & \textcolor{red}{More (0.86)} & \textcolor{red}{More (0.94)} & \textcolor{red}{More (0.93)}\\
$q_{trans}$ &If the weather has a lot of snow, will not the squirrels gain weight? [$a^*_{trans}$: \textcolor{red}{More}] & \textcolor{blue}{Less (0.75)} & \textcolor{red}{More (0.43)} & \textcolor{red}{More (0.87)}\\
\bottomrule
\end{tabular}
\begin{tabular}{C{0.85cm} | L{9.5cm}  |C{1.85cm} C{1.85cm} }
& HotpotQA (comparison) Input & RoBERTa & \texttt{DA} \\
\midrule
$p$& B. Reeves Eason is a film director, actor and screenwriter. Albert S. Rogell a film director. && \\
$q$& Who has more scope of profession, B. Reeves Eason or Albert S. Rogell? [$a^*$: \textcolor{red}{B. Reeves Eason}] & \textcolor{red}{B. Reeves Eason} & \textcolor{red}{B. Reeves Eason}\\
$q_{sym}$ & Who has less scope of profession, B. Reeves or Albert S. Rogell? [$a_{sym}^{*}$: \textcolor{blue}{Albert S. Rogell}] & \textcolor{red}{B. Reeves Eason} & \textcolor{blue}{Albert S. Rogell} \\\bottomrule
\end{tabular}
}
\end{center}
\caption{\label{tab:output_example} Qualitative comparison of RoBERTa, + \texttt{DA},  + \texttt{DA} + \texttt{Reg}. 
The examples are partially modified.
}
\end{table*}

We use standard F1 and EM scores for performance evaluation on HotpotQA and use accuracy for WIQA and QuaRel. 
We report a violation of consistency following~\citet{minervini-riedel-2018-adversarially} to evaluate the effectiveness of our approach for improving prediction consistencies. 
We compute the violation of consistency metric $v$ as the percentage of examples that do not agree with symmetric  and transitive logical rules. 
More model and experimental details are in Appendix.

\vspace{-.2cm}
\paragraph{Main Results}
\Tref{tab:result_wiqa} demonstrates that our methods (\texttt{DA} and \texttt{DA} + \texttt{Reg}) constantly give 1 to 5 points improvements over the state-of-the-art RoBERTa QA's performance on all three of the datasets, advancing the state-of-the-art scores on WIQA and QuaRel by 4.7\% and 8.4\%, respectively. 
On all three datasets, our method significantly reduces the inconsistencies in predictions, demonstrating the effects of both data augmentation and regularization components. Notably on WIQA, RoBERTa shows violation of consistency in 13.9\% of the symmetric examples and 10.0\% of the transitive examples. Our approach reduces the violations of symmetric and transitive consistencies to 8.3\% and 2.5\%, respectively.

\paragraph{Results with limited training data}
Table~\ref{tab:result_wiqa} also shows that our approach is especially effective under the scarce training data setting: when only 20\% of labeled data is available, our \texttt{DA} and \texttt{Reg} together gives more than 12\% and 14\% absolute accuracy improvements over the RoBERTa baselines on WIQA and QuaRel, respectively. 

\paragraph{Ablation study}
We analyze the effectiveness of each component on \Tref{tab:ablation_analysis}. 
\texttt{DA} and \texttt{Reg} each improves the baselines, and the combination performs the best on WIQA and QuaRel.  
\texttt{DA} (standard) follows a previous standard data augmentation technique that paraphrases words (verbs and adjectives) using linguistic knowledge, namely WordNet~\cite{miller1995wordnet}, and does not incorporate logical rules.
Importantly, \texttt{DA} (standard) does not give notable improvement over the baseline model both in accuracy and consistency, which suggests that logic-guided augmentation gives additional inductive bias for consistent QA beyond amplifying the number of train data.
As WIQA consists of some transitive or symmetric examples, we also report the performance with \texttt{Reg} only on WIQA. The performance improvements is smaller, demonstrating the importance of combining with \texttt{DA}.

\paragraph{Qualitative Analysis}

Table~\ref{tab:output_example} shows qualitative examples, comparing our method with RoBERTa baseline. 
Our qualitative analysis shows that \texttt{DA+Reg} reduces the confusion between opposite choices, and assigns larger probabilities to the ground-truth labels for the questions where \texttt{DA} shows relatively small probability differences. 

On HotpotQA, the baseline model shows large consistency violations as shown in \Tref{tab:result_wiqa}.
The HotpotQA example in Table~\ref{tab:output_example} shows that RoBERTa selects the same answer to both $q$ and $q_{sym}$, while 
\texttt{DA} answers correctly to both questions, demonstrating its robustness to surface variations. 
We hypothesize that the baseline model exploits statistical pattern, or dataset bias presented in questions and that our method reduces the model's tendency to exploit those spurious statistical patterns~\cite{he2019unlearn,elkahky-etal-2018-challenge}, which leads to large improvements in consistency.

\section{Conclusion}We introduce a logic guided data augmentation and consistency-based regularization framework for accurate and globally consistent QA, especially under limited training data setting. Our approach significantly improves the state-of-the-art models across three substantially different QA datasets. Notably, our approach advances the state-of-the-art on QuaRel and WIQA, two standard benchmarks requiring rich logical and language understanding. 
We further show that our approach can effectively learn from extremely limited training data. 

\section*{Acknowledgments}
This research was supported by ONR N00014-18-1-2826, DARPA N66001-19-2-403, NSF (IIS1616112, IIS1252835), Allen Distinguished Investigator Award, Sloan Fellowship, and The Nakajima Foundation Fellowship. We thank Antoine Bosselut, Tim Dettmers, Rik Koncel-Kedziorski, Sewon Min, Keisuke Sakaguchi, David Wadden, Yizhong
Wang, the members of UW NLP group and AI2, and the anonymous reviewers for their insightful  feedback.

\bibliography{acl2020}
\bibliographystyle{acl_natbib}

\newpage
\appendix
\section{Details of Human Annotations}
In this section, we present the details of human annotations used for symmetric example creation (the (a) operation).
We first sample the most frequent 500 verbs, 50 verb phrases and 500 adjectives from from the WIQA and QuaRel training data. 
Then, human annotators select words with some polarity (e.g., increase, earlier). 
Subsequently, they annotate the antonyms for each of the selected verbs and adjectives. 
Consequently, we create 64 antonym pairs mined from a comparison QA dataset. 
We reuse the same dictionary for all three datasets. Examples of annotated antonym pairs are shown in Table~\ref{tab:antonym_table}. 

\begin{table}[h!]
    \centering
    \small
\begin{tabular}{l  ?  l}\toprule
adjectives & verbs \& verb phrases \\ \midrule
more $\leftrightarrow$ less & increase $\leftrightarrow$  decrease \\
slowly $\leftrightarrow$ quickly & heat up $\leftrightarrow$  cool down   \\
stronger $\leftrightarrow$ weaker & lose weight $\leftrightarrow$ gain weight\\
later $\leftrightarrow$ earlier & raise $\leftrightarrow$ drop \\
younger $\leftrightarrow$ older& remove $\leftrightarrow$ add \\ 

\bottomrule
\end{tabular}
\caption{Ten examples of annotated antonyms for comparison type questions. }\label{tab:antonym_table}
\end{table}

\section{Details of answer re-labeling on WIQA and HotpotQA}
We present the details of answer re-labeling operations in WIQA and HotpotQA, where the number of the answer candidates is more than two. 

\paragraph{Answer re-labeling in WIQA (symmetric)}
In WIQA, each labeled answer $a^{*}$ takes one of the following values: \{{\it more}, {\it less}, {\it no effects}\}. Although {\it more} and {\it less} are opposite, {\it no effects} is a neutral choice. 
In addition, in WIQA, a question $q$ consists of a cause $c$ and an effect $e$, and we can operate the three operations (a) replacement, (b) addition and (c) removal of words. 
When we add the operations to both of $c$ and $e$, it would convert the question to opposite twice, and thus the original answer remains same. When we add one of the operation to either of $c$ or $e$, it would convert the question once, and thus, the answer should be the opposite one. 
Given these two assumption, we re-label answer as: (i) if we apply only one operation to either $e$ or $c$ and $a^{*}$ is {\it more} or {\it less}, the $a_{sym}^{*}$ will be the opposite of $a^{*}$, 
(ii) if we apply only one operation to either $e$ or $c$ and $a^{*}$ is {\it no effect}, the $a_{sym}^{*}$ will remain  {\it no effect}, and (iii) if we apply one operation to each of $e$ and $c$, the $a_{sym}$ remains the same.

\paragraph{Answer re-labeling in WIQA (transitive)}
For transitive examples, we re-label answers based on two assumptions on causal relationship. 
A transitive questions are created from two questions, $X_1 = (q_1, p, a_1^{*})$ and $X_2 = (q_2, p, a_2^{*})$, where $q_1$ and $q_2$ consist of $(c_1, e_1)$ and $(c_2, e_2)$ and $e_1=c_2$ holds. 
If $a_1$ for $X_1$ is ``more'', it means that the $c_1$ causes $e_1$. 
$e_1$ is equivalent to the cause for the second question ($c_2$), and $a_2^{*}$ represents the causal relationship between $c_2$ and $e_2$. 
Therefore, if $a_1^{*}$ is a positive causal relationship, $c_1$ and $e_2$ have the relationship defined as $a_2^{*}$.
We assume that if the $a_1^{*}$ is ``more'', $a_3^{*}(=a_{trans}^{*})$ will be same as $a_2$, and re-label answer following this assumption. 


\paragraph{Answer re-labeling in HotpotQA}
In HotpotQA, answer candidates $\mathbf{A}$ are not given. Therefore, we extract possible answers from $q$. 
We extract two entities included in $q$ by string matching with the titles of the paragraphs given by the dataset. 
If we find two entities to be compared and both of them are included in the gold paragraphs, we assume the two entities are possible answer candidates. The new answer $a_{sym}^{*}$ will be determined as the one which is not the original answer $a^{*}$.

\section{Details of Baseline Models}
We use RoBERTa~\cite{li-etal-logical-nli-2019} as our baseline. 
Here, we present model details for each of the three different QA datasets.

\paragraph{Classification-based model for WIQA}
As the answer candidates for WIQA questions are set to \{more, less, no effects\}, we use a classification based models as studied for NLI tasks. 
The input for this model is $\texttt{[CLS]}$ $p$ $\texttt{[SEP]}$ $q$ $\texttt{[SEP]}$. We use the final hidden vector corresponding to the first input token ($\texttt{[CLS]}$) as the aggregate representation. 
We then predict the probabilities of an answer being a class $C$ in the same manner as in \cite{devlin-etal-2019-bert,liu2019roberta}. 

\paragraph{Multiple-choice QA model for QuaRel}
For QuaRel, two answer choices are given, and thus we formulate the task as multiple-choice QA. 
In the original dataset, all of the $p$, $q$ and $\mathbf{A}$ are combined together (e.g., {\it The fastest land animal on earth, a cheetah was having a 100m race against a rabbit. Which one won the race? (A) the cheetah (B) the rabbit}), and thus we process the given combined questions into $p$, $q$ and $\mathbf{A}$ (e.g., the question written above will be $p=${\it The fastest land animal on earth, a cheetah was having a 100m race against a rabbit. }, $q=${Which one won the race?} and $\mathbf{A}=$\{the cheetah, rabbit\}).
Then the input will be $\texttt{[CLS]}$ $p$ $\texttt{[SEP]}$ {\it ``Q: ''} $q$~{\it ``A: ''} $a_i$ $\texttt{[SEP]}$, and we will use the final hidden vector corresponding to the first input token ($\texttt{[CLS]}$) as the aggregate representation.
We then predict the probabilities of an answer being an answer choice $a_i$ in the same manner as in \cite{liu2019roberta}. 

\paragraph{Span QA model for HotpotQA}
We use the RoBERTa span QA model studied for SQuAD~\cite{devlin-etal-2019-bert,liu2019roberta} for HotpotQA. As we only consider the questions whose answers can be extracted from $p$, we do not add any modifications to the model unlike some previous studies in HotpotQA~\cite{min2019multi,cognitive_graph_2019}. 

\section{Details of Implementations and Experiments }
\paragraph{Implementations}
Our implementations are all based on PyTorch. 
In particular, to implement our classification based and span-based model, we use \texttt{pytorch-transformers}~\cite{wolf2019transformers}\footnote{\url{https://github.com/huggingface/transformers}}. To implement our multiple choice model, we use \texttt{fairseq}~\cite{ott2019fairseq}\footnote{\url{https://github.com/pytorch/fairseq}}. 

\paragraph{Hyper-parameters}
For HotpotQA, we train a model for six epochs in total. 
For the model without data augmentation or regularization, we train on the original dataset for six epochs. 
For the models with data augmentation, we first train them on the original HotpotQA train data (including both bridge and comparison questions) for three epochs, and then train our model with augmented data and regularization for three epochs. 
For HotpotQA, we train our model with both bridge and comparison questions, and evaluate on comparison questions whose answers can be extracted from the context. 

Due to the high variance of the performance in the early stages of the training for small datasets such as QuaRel or WIQA, for these two datasets, we set the maximum number of training epochs to 150 and 15, respectively. 
We terminate the training when we do not observe any performance improvements on the development set for 5 epochs for WIQA and 10 epochs for QuaRel, respectively. 
We use Adam as an optimizer ($\epsilon = 1{\rm E}-8$) for all of the datasets. 
Other hyper-parameters can be seen from Table~\ref{tab:hyper_parameters}
\begin{table}[h!]
    \centering
    \small
\begin{tabular}{l ? c c c }\toprule
hyper-parameters & WIQA & QuaRel & HotpotQA  \\ \midrule
train batch size & 4 & 16 & 12 \\
gradient accumulation & 16  & 1 & 1 \\
max token length & 256 & 512 & 384 \\
doc stride & -- & -- & 128 \\
learning rate & 2E-5 & 1E-5 & 5E-5\\
weight decay & 0.01 & 0.01 & 0.0 \\
dropout &0.1 & 0.1 & 0.1\\
warm up steps & 0 & 150 & 0 \\
$\tau$ for annealing & 3 & 25 & 3 \\
$\lambda_{sym}$ & 0.5  & 0.1 & 0.25 \\
$\lambda_{trans}$  & 0.05 & -- & -- \\
\bottomrule
\end{tabular}
\caption{Ten examples of annotated antonyms for comparison type questions. }\label{tab:hyper_parameters}
\end{table}

\section{Qualitative Examples on HotpotQA}
As shown in Table 2, the state-of-the-art RoBERTa model produces a lot of consistency violations. 
Here, we present several examples where our competitive baseline model cannot answer correctly, while our RoBERTa+DA model answers correctly. 

\paragraph{A question requiring world knowledge}
One comparison question asks ``Who has \textcolor{red}{\bf more} scope of profession, B. Reeves Eason or Albert S. Rogell'', given context that B. Reeves is an American film director, actor and screenwriter and Albert S. Rogell is an American film director. The model correctly predicts ``B. Reeves Eason'' but fails to answer correctly to  ``Who has \textcolor{blue}{\bf less} scope of profession, B. Reeves Eason or Albert S. Rogell'', although the two questions are semantically equivalent. 

\paragraph{A question with negation}
We found that due to this reasoning pattern our model struggles on questions involving negation.
Here we show one example. 
We create a question by adding a negation word, $q_{sym}$,``Which species is \hl{not} native to asia, corokia or rhodotypos?'', where we add negation word \hl{not} and the paragraph corresponding to the question is $p=$``{\bf Corokia} is a genus in the Argophyllaceae family comprising about ten species native to New Zealand and one native to Australia. {\bf Rhodotypos scandens} is a deciduous shrub in the family Rosaceae and is native to China, possibly also Japan.''. 
The model predicts {\bf Rhodotypos scandens}, while the model predicts the same answer to the original question $q$, `which species is native to asia, corokia or rhodotypos?''. 
This example shows that the model strongly relies on surface matching (i.e., ``native to'') to answer the question, without understanding the rich linguistic phenomena or having world knowledge. 

\end{document}